\documentclass[letterpaper]{article} 
\usepackage[draft]{aaai25}  
\usepackage{times}  
\usepackage{helvet}  
\usepackage{courier}  
\usepackage[hyphens]{url}  
\usepackage{graphicx} 
\usepackage{natbib}  
\usepackage{caption} 
\usepackage{algorithm}
\usepackage{amsmath}
\usepackage{float}
\usepackage{caption}
\usepackage{subfig}
\usepackage{color}
\usepackage{multirow}
\usepackage{cite}
\usepackage{booktabs}
\usepackage{array}
\usepackage{xcolor}
\usepackage{algpseudocode}
\usepackage{multirow,bm}
\usepackage{cleveref}
\usepackage{amssymb}
\usepackage{mathtools}
\usepackage{newfloat}
\usepackage{listings}
\DeclareCaptionStyle{ruled}{labelfont=normalfont,labelsep=colon,strut=off} 
\floatstyle{ruled}
\newfloat{listing}{tb}{lst}{}
\floatname{listing}{Listing}
\title{Enhancing Solutions for Complex PDEs: Introducing Complementary Convolution and Equivariant Attention in Fourier Neural Operators}
\author{
    Xuanle Zhao\textsuperscript{\rm 1,2}\equalcontrib,
    Yue Sun\textsuperscript{\rm 1}\equalcontrib,
    Tielin Zhang\textsuperscript{\rm 1,2},
    Bo Xu\textsuperscript{\rm 1,2}
}
\affiliations{
    \textsuperscript{\rm 1}Institute of Automation, Chinese Academy of Sciences \\
    \textsuperscript{\rm 2}School of Artificial Intelligence, University of Chinese Academy of Sciences \\

}

\usepackage{bibentry}

\begin{document}

\maketitle

\begin{abstract}
Neural operators improve conventional neural networks by expanding their capabilities of functional mappings between different function spaces to solve partial differential equations (PDEs).  One of the most notable methods is the Fourier Neural Operator (FNO), which draws inspiration from Green's function method and directly approximates operator kernels in the frequency domain. However, after empirical observation followed by theoretical validation, we demonstrate that the FNO approximates kernels primarily in a relatively low-frequency domain.  
This suggests a limited capability in solving complex PDEs, particularly those characterized by rapid coefficient changes and oscillations in the solution space. Such cases are crucial in specific scenarios, like atmospheric convection and ocean circulation.
To address this challenge, inspired by the translation equivariant of the convolution kernel, we propose a novel hierarchical Fourier neural operator along with convolution-residual layers and attention mechanisms to make them complementary in the frequency domain to solve complex PDEs. We perform experiments on forward and reverse problems of multiscale elliptic equations, Navier-Stokes equations, and other physical scenarios, and find that the proposed method achieves superior performance in these PDE benchmarks, especially for equations characterized by rapid coefficient variations.
\end{abstract}

\section{Introduction}
\label{sec:intro}
Conventional research in the deep learning field is predominantly driven by neural networks designed to learn relationships between input and output pairs that lie in finite-dimensional spaces. These designs have led to notable breakthroughs in numerous domains, including computer vision \cite{he2016deep, chen2018neural}, natural language processing \cite{zhou2016attention, devlin2018bert}, reinforcement learning \cite{hafner2019dream}, and signal processing \cite{dong2018speech, gulati2020conformer}.

Partial differential equations (PDEs) are utilized in scientific research to describe a wide variety of physical, chemical, and biological phenomena \cite{sommerfeld1949partial, zhang2023brain}. From turbulent flows to atmospheric circulation and material stress analysis, numerous real-world phenomena are fundamentally governed by the underlying PDEs. Therefore, solving PDEs is a crucial aspect of addressing problems in natural science.

Traditional numerical methods for solving PDEs, such as the finite element method (FEM) and the finite difference method (FDM), present challenges in terms of incorporating noisy data, generating complex meshes, solving high-dimensional problems, and handling inverse problems. 
Fortunately, by harnessing the expressiveness of neural networks, many innovative methods \cite{hao2023gnot, brandstetter2022clifford, liu2023nuno, li2022transformer, zhao2024ode} have been proposed to effectively overcome the limitations of numerical methods in solving PDEs.
Examples of such methods include physics-informed neural networks (PINN) \cite{karniadakis2021physics} and Galerkin transformers (GT) \cite{cao2021choose}, which are specifically designed neural networks for PDE simulation by estimating the mapping between inputs and output pairs. In addition, methods such as the deep operator network (DeepONet) \cite{lu2019deeponet} and the Fourier Neural Operator (FNO) \cite{li2020fourier, li2022fourier} attempt to learn an operator between input and output function spaces. Besides solving PDEs, these methods have also proven to be effective in dealing with issues related to complex dynamics such as climate change and natural disasters \cite{gopakumar2023fourier, pathak2022fourcastnet, zhao2024enhanced}.

We preferentially consider more complex PDEs, such as multiscale PDEs, which have been widely used in physics, engineering, and related disciplines for analyzing complex practical problems such as reservoir modeling, ocean circulation, and high-frequency scattering \cite{quarteroni2003analysis}. Given that complex PDEs are characterized by rapidly changing coefficients and oscillations in the solution space, it is crucial to capture information across various scales and frequency ranges. Nevertheless, when applied to solving PDEs, evidence shows that FNO and related methods tend to prioritize learning low-frequency components \cite{li2020fourier, liu2022ht, anonymous2023dilated}, which raises the key question about how to capture high-frequency features and combine them with low-frequency features of the FNO for solving complex PDEs.

In this paper, we presented a novel attentive hierarchical FNO method that attempts to capture and integrate low-frequency and high-frequency features at different scales.
Since the convolution kernel is locally computed, high-frequency local details can be captured efficiently. 
Therefore, inspired by DCNO \cite{anonymous2023dilated}, we adopt the Fourier kernel with the convolutional-residual layer, which aims to improve the ability to capture high-frequency information.
Furthermore, we theoretically prove that the attention mechanism composed of channel and spatial attention could integrate before the Fourier layer in a translation equivariant way. Furthermore, to effectively capture information across different scales, we propose a hierarchical architecture that learns convolutional-residual Fourier layers and equivariant attentions at multiple scales.
Our main contributions can be summarized as follows:
\begin{itemize}
    \item We propose an enhanced method to address the issue where FNO-related approaches struggle to capture high-frequency features effectively. Specifically, our method integrates high-frequency and low-frequency components simultaneously with convolutional-residual Fourier layers and equivariant attentions in a hierarchical structure.
    \item The proposed method surpasses previous state-of-the-art approaches in existing PDE benchmarks, including Navier-Stokes equations, Darcy equations, and particularly multiscale elliptic equations with rapidly changing coefficients and significant solution variations.
    \item Furthermore, our method demonstrates effectiveness and robustness in solving inverse PDE problems, particularly when dealing with noisy input data.
\end{itemize}

\section{Related works}
We briefly cover the background and related works in this section. More related works are listed in the Appendix.

\subsection{Neural PDE solver}
Many excellent algorithms have been proposed previously for solving PDEs using neural networks \cite{long2018pde, hao2022physics}. Physics-informed neural network (PINN) \cite{karniadakis2021physics} incorporates PDEs into the network by giving additional constraints with the form of PDEs into loss function, which then guide the synaptic modifications towards tuned parameters that satisfy data distribution, physical PDE laws, and other necessary boundary conditions. GT \cite{cao2021choose} utilizes the self-attention mechanism to build operator learners to solve PDEs and designs Fourier-type and Galerkin-type attention with linear complexity to reduce the computation cost.
Neural operators leverage the concept that the operator denotes the mapping between infinite input and output function spaces. 
DeepONet \cite{lu2019deeponet} leverages the universal approximation theorem to derive a branch-trunk structure to form the operator in a polynomial regression way. Some other methods incorporate trained neural networks into conventional numerical solvers, to minimize numerical errors when dealing with coarse grids \cite{cuomo2022scientific, meng2020ppinn}. Our approach is also built on the concept of FNO \cite{li2020fourier, li2022fourier}, which utilizes Green's function and directly approximates the kernel in the Frequency domain. We incorporate different Fourier features at different frequency scales, which as a result, reaches promising performance in solving complex PDEs (e.g., multiscale PDEs).

\subsection{Multiscale PDEs}
Multiscale PDEs have many applications, including forecasting atmospheric convection and ocean circulation, modeling the subsurface of flow pressures \cite{furman2008modeling, huyakorn2012computational}, the deformation of elastic materials \cite{rivlin1948large, merodio2003instabilities}, and the electric potential of conductive materials \cite{sundnes2005operator}. Multiscale elliptic PDEs are classic examples of multiscale PDEs. 
Solving elliptic PDEs with smooth coefficients is a conventional problem that can be effectively addressed using FNO. However, when the coefficients become non-smooth and exhibit rapidly changing features, the values in the solution spaces can exhibit oscillations and high contrast ratios \cite{anonymous2023dilated}. Another example is turbulent flow, which is modeled by the Navier-Stokes equation. This equation describes fluid dynamics and exhibits turbulent behavior at high Reynolds numbers. In turbulent flow, unsteady vortices interact, leading to complex dynamics. To solve these multiscale PDEs effectively, models must account for both global and local dynamics.
Equations and more details about multiscale PDEs are presented in the Appendix.

\subsection{Fourier Neural Operator}
Fourier neural operator (FNO) \cite{li2020fourier, li2022fourier} draws inspiration from the conventional Green's function method and directly optimizes the kernel within the Fourier frequency domain by utilizing the Fast Fourier Transform (FFT). This approach has been demonstrated to be an efficient means of reducing computational cost and performing global convolution.
A notable advancement is that the operator kernel is directly trained in the frequency domain, whereby the network is theoretically independent of the training data resolution. Thus, FNO \cite{li2020fourier, li2022fourier} can deal with super-resolution problems and be trained on multiple PDEs.

FNO has laid the foundation for operator learning, inspiring several subsequent works in the field. Geo-FNO \cite{li2022fourier} deforms the irregular grid into a latent space with a uniform grid to solve the limitation of FFT which could only be applied to rectangular domains. F-FNO \cite{tran2021factorized} learns the kernel weights in a factorized way with separable spectral layers. G-FNO \cite{helwig2023group} utilizes the symmetry groups in the Fourier kernel to learn equivalent representations and improve accuracy even under imperfect symmetries.

However, FNO ignores high-frequency components by default to learn a smooth representation of the input space, which results in poor performance when solving PDEs with rapidly changing coefficients \cite{liu2022ht}. To address this limitation, we integrate convolutional-residual Fourier layers and equivariant attention mechanisms to capture local and global frequency features simultaneously.

\begin{figure*}[htbp]
    \centering
    \includegraphics[width=0.95\linewidth]{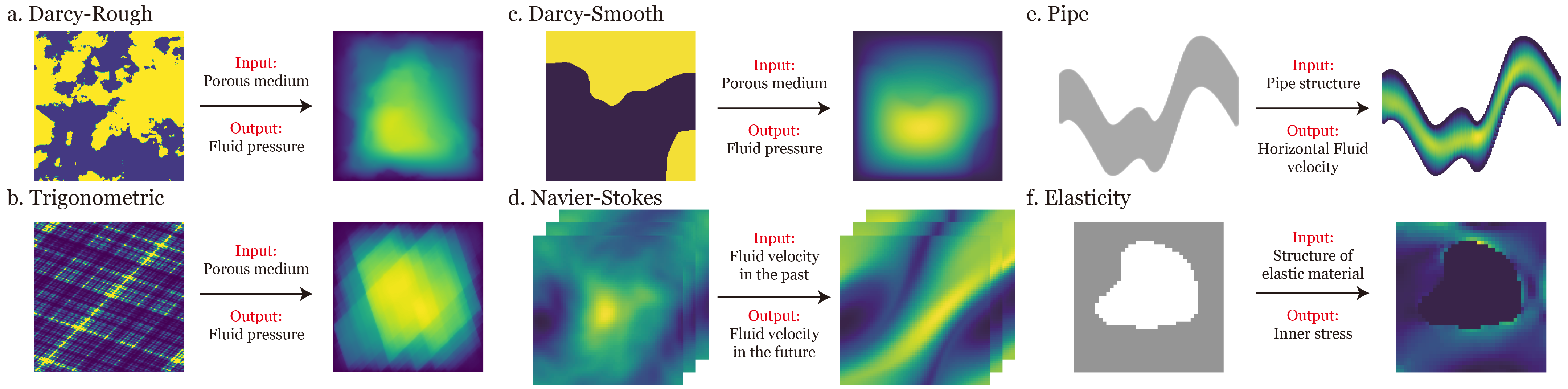}
    \caption{Examples of various tasks, including Darcy-Rough, Trigonometric, Darcy-Smooth, Navier-Stokes, Pipe, and Elasticity datasets. These datasets solve equations according to coefficients, previous solutions, and structures by approximating mappings between input and output in coordinate spaces. All these tasks are covered in experimental verification.}
    \label{fig2}
    
\end{figure*}

\section{Methods}
In this section, we introduce our proposed method in detail. First, we formulate the definition of operator learning and illustrate FNO only learning the low-frequency domain. Then, we theoretically prove the theorem to combine the attention mechanism and Fourier transformation in a translation-equivariant way. Finally, we introduce the architecture of our proposed model.

\subsection{Problem Formulation}
For a PDE problem, the observed samples are ${(a_i, u_i)}_{i=1}^N$. Assuming the coordinates in a bounded open set $\mathcal{D}\subset\mathbb{R}^d$, both the input and output can be expressed as functions with respect to these coordinates. These functions belong to the Banach spaces $\mathcal{X}=\mathcal{X}(\mathcal{D};\mathbb{R}^{d_{\boldsymbol{a}}})$ and $\mathcal{Y}=\mathcal{Y}(\mathcal{D};\mathbb{R}^{d_{\boldsymbol{u}}})$ respectively. Here, $\mathbb{R}^{d_{\boldsymbol{a}}}$ and $\mathbb{R}^{d_{\boldsymbol{u}}}$ denotes the range of input and output functions. \textbf{$\mathcal{D}$} consists of a finite set of grid points within a rectangular area in $\mathbb{R}^2$. The function values are represented by position $x\subset\mathcal{D}$, which could be denoted as $a(x)$ and $u(x)$.
The overall solving process could be viewed as using a neural network $f_\theta$ to approximate the mapping $\mathcal{X}\to\mathcal{Y}$ to predict the output $\hat{u}(x)$, where $\hat{u}(x)=f_\theta(a)(x)$.

The Fourier neural operator \cite{li2020fourier} is a powerful and efficient architecture for modeling PDEs that learn operators for mapping input and output function spaces. This algorithm is inspired by Green's function method by learning the kernel integral operator defined below. 
 
Define the kernel integral operator mapping  by
\begin{equation}\label{KIO}
    [\mathcal{K}(\phi) a](x)=\int k_\phi(x,y)a(y)dy,\quad \forall x \in \mathcal{D},
\end{equation}
where Green's function $k_\phi $  is parameterized by neural networks with parameter $\phi$.
In this context, the function $k_{\phi}$ serves as a kernel function that is learned from data. FNO assumes that Green’s function is periodic and only dependent on the relative distance, which means that $k_{\phi}(x,y)=k_{\phi}(x-y)$. Then the operation in Eq. \ref{KIO} could be regarded as convolution and efficiently implemented as element-wise multiplication in the frequency domain by using the convolution theorem:
\begin{equation}
    \begin{aligned}
    [\mathcal{K}(\phi) a](x)=&\int k_{\phi}(x- y)a(y)dy \\
    =&\mathcal{F}^{-1}\left(\mathcal{F}(k_{\phi})\cdot \mathcal{F}(a)\right)(x) \\
    =&\mathcal{F}^{-1}\left(R_{\phi}\cdot \mathcal{F}(a)\right)(x) ,      
    \end{aligned}
\end{equation}
where $\mathcal{F}$ and $\mathcal{F}^{-1}$ are the Fourier transform and the inverse Fourier transform, respectively. Instead of learning the kernel $k_{\phi}$, FNO directly learns the kernel $R_{\phi}$ in the Fourier domain.

It is important to note that although the integral operator itself is linear, the neural operator can learn non-linear operators by using non-linear activation functions. Thus, the Fourier layer can be formally expressed as:
 \begin{equation}\label{fno_layers}
     \hat{u}(x):=\sigma\left(W a(x)+\mathcal{F}^{-1}(R_{\phi}\cdot \mathcal{F}(a))(x)\right), \ \forall x \in \mathcal{D},
 \end{equation}
where  $\sigma$ denotes the non-linear activation function, $W$ and $R_{\phi}$ are the fully connected layer and trainable operator kernel, respectively.

\subsection{FNO Drawbacks}
Nevertheless, we take a one-dimensional case as an example to show that high-frequency features are not well represented in FNO and related methods, posing a challenge in dealing with multiscale PDEs. 
During the Fourier transformation process of the FNO, only the low-frequency components ($\omega\leq T_\omega$) are reserved for multiplication, and high-frequency components ($\omega> T_\omega$) are ignored by default. The size of the kernel is the same as the size of the reserved low-frequency components. Thus the elementwise multiplication process could be expressed as:
\begin{equation*}
    (R_{\phi}\cdot\mathcal{F}(a))(\omega) = \left\{
	\begin{aligned}
		&(R_{\phi}\cdot\mathcal{F}(a))(\omega), \ &\omega\leq T_\omega, \\
		&0,        \ &\omega>T_\omega.
	\end{aligned}
    \right. 
\end{equation*}
Therefore, after inverse Fourier transformation,  
\begin{equation*}
  \begin{aligned}
    \mathcal{F}^{-1}(R_{\phi}\cdot \mathcal{F}(a))(x)=\sum_{\omega\leq T_\omega} (R_{\phi}\cdot\mathcal{F}(a))(\omega)e^{i\omega x},
\end{aligned}  
\end{equation*}
only the low-frequency components are represented. 
For the sake of notational convenience, we only use the one-dimensional case for illustration, which still stands for 2D and 3D cases. Additionally, previous works \cite{liu2022ht} also inform that FNO and GT have shown their tendency to prioritize learning low-frequency components before high-frequency components when applied to multiscale PDEs. Therefore, obtaining information about different input function frequencies is crucial to solving complex PDEs. Motivated by the need to handle PDEs at multiple scales, this paper introduces a novel hierarchical attentive Fourier neural operator. 

\subsection{Attentive Equivariant Convolution}
In the FNO \cite{li2020fourier}, the kernel of Green's function is imposed as the convolution kernel, which is a natural choice from the perspective of fundamental solutions. A fundamental property of the convolution is that it commutes with translations, 
\begin{equation}
\mathcal{L}_y\left[f \star \psi\right](x)=\left[\mathcal{L}_y[f] \star \psi\right](x)
\end{equation}
where $L_y$ is the translation operator.\footnote{It follows that $\mathcal{L}_g[f](x)=f\left(g^{-1} x\right)=f(x-y)$, where $g^{-1}=-y$ is the inverse of $g$ in the translation group $\left(\mathbb{R}^d,+\right)$ for $g =y$.} In other words, convolving a $y$-translated signal $\mathcal{L}_y[f]$ with a filter is equivalent to first convolving the original signal $f$ with the filter $\psi$ and $y$-translating the obtained response next. This property is referred to as translation equivariance.

\begin{figure*}[tbp]
    \centering
    \includegraphics[width=0.98\linewidth]{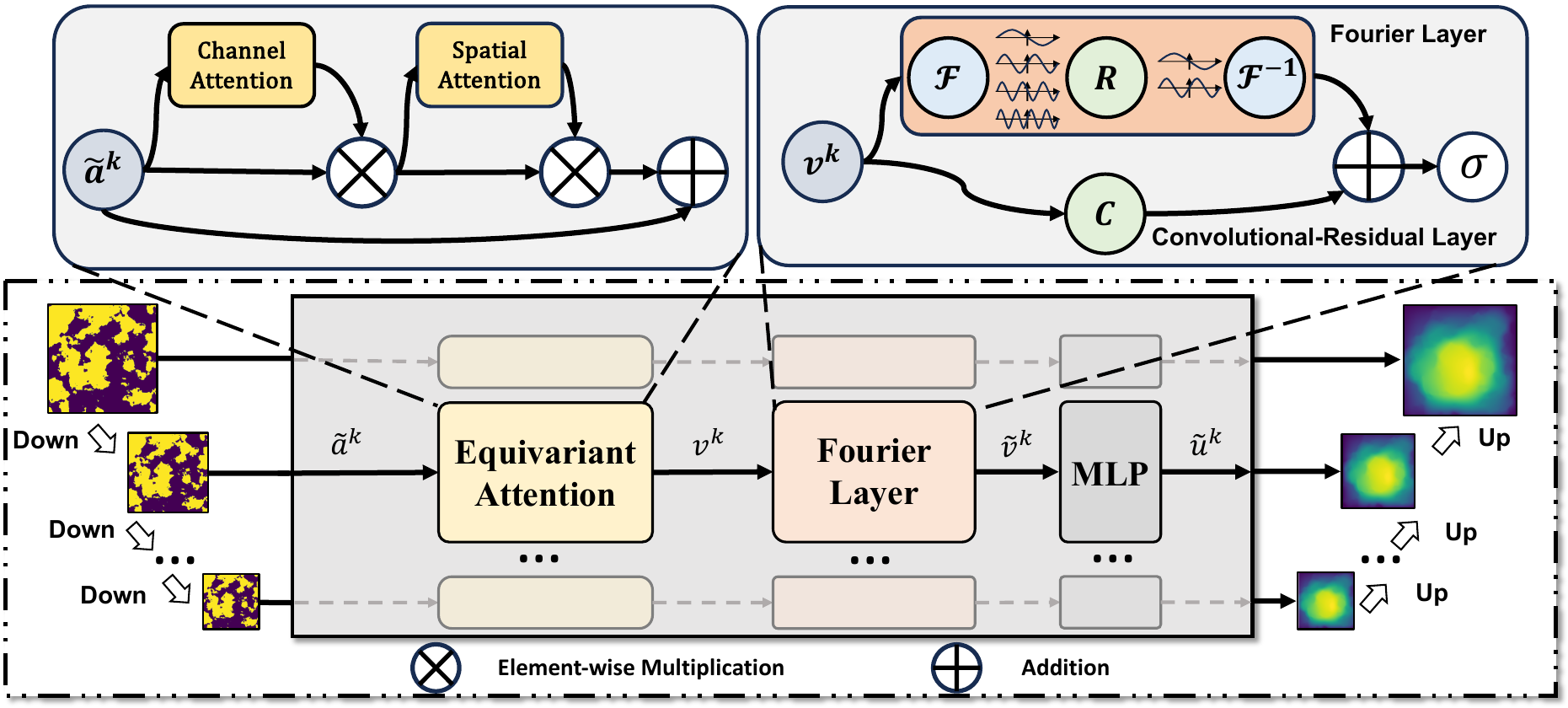}
    \caption{The overall network architecture. The input is downsampled and processed at each scale using equivariant attention and convolutional-residual Fourier layers. The final output is obtained by upsampling the outputs of hierarchical layers.}
    \label{fig3}
    \vspace{-5pt}
\end{figure*}

Previous works have defined attentive group convolution \cite{romero2020attentive, cohen2016group} and proved its equivariant property. We simplify them into attentive convolution defined on $\mathbb{R}^d$,
\begin{equation}\label{attentive_convlution}
\left[f \star^\alpha \psi\right](x)= \int_{\mathbb{R}^d} \alpha(x, \tilde{x}) f(\tilde{x}) \mathcal{L}_x\left[\psi\right](\tilde{x}) \mathrm{d} \tilde{x}
\end{equation}
where $\alpha(x, \tilde{x})$ is the attention map between the input and output positions.
\paragraph{\textbf{Theorem 1.}} \textit{The attentive convolution is an equivariant operator if and only if the attention operator $\mathcal{A}$ satisfies:
\begin{equation}\label{thm01}
\forall_{\bar{x}, x, \tilde{x} \in \mathbb{R}^d}: \mathcal{A}\left[\mathcal{L}_{\bar{x}} f\right](x, \tilde{x})=\mathcal{A}[f]\left(\bar{x}^{-1} x, \bar{x}^{-1} \tilde{x}\right)
\end{equation}
If, moreover, the maps generated by $\mathcal{A}$ are invariant to one of its arguments, and, hence, exclusively attend to either the
input or the output domain, then $\mathcal{A}$ satisfies \Cref{thm01} if it is equivariant and thus, based on convolutions.}

Since convolutions and pooling operations are translation equivariant, mostly visual attention mechanisms are translation equivariant as well \cite{romero2020attentive}. One special case is channel attention based on fully connected layers (a non-translation equivariant map) in SE-Nets \cite{hu2018squeeze}. However, the input of the fully connected layers is obtained via global average pooling, which has shown that it is equivalent to a pointwise convolution \cite{romero2020attentive}. Therefore, attention here is translation equivariant \cite{cohen2018spherical}.

Furthermore, previous works broadly assumed that the maps in visual attention do not depend on the filters $\psi$ and could be equivariantly factorized into spatial $\alpha^{\mathcal{X}}$ and channel $\alpha^{\mathcal{C}}$ components. Hence, the attention coefficient $\alpha$ is the sole function of the input signal and becomes only
dependent on $x$ (\Cref{attentive_convlution}). 
\begin{equation}\label{attconv}
\left[f \star^\alpha \psi\right]=\left[f^\alpha \star \psi\right]=\left[\left(\alpha f\right) \star \psi\right]=\left[\left(\alpha^{\mathcal{X}} \alpha^{\mathcal{C}} f\right) \star \psi\right]
\end{equation}
In this way, the attention maps $\alpha$ can be shifted to the input feature map $f$.
Resultantly, the attentive convolution is reduced to a sequence of conventional convolutions and point-wise non-linearities (Thm. 1), which further reduces the computational cost of attention.

Inspired by GFNO \cite{helwig2023group}, we further utilize the following group Fourier transformation theorem to illustrate our Attentive Equivariant Fourier Neural Operator. 
\paragraph{\textbf{Theorem 2.}} \textit{Given the orthogonal group O(d) acting on functions defined on $\mathbb{R}^d$ by the map $(g, f) \mapsto L_g f$ where $\left(L_g f\right)(x):=f\left(g^{-1} x\right)$, the group action commutes with the Fourier-transform, i.e. $\mathcal{F} \circ L_q=L_q \circ \mathcal{F}$.}

This theorem describes the equivariance of the Fourier transformation, which means applying a transformation from $O(d)$ to a function in physical space equally applies the transformation to the Fourier transform of the function. Hence, based on \Cref{fno_layers}, our model could be generally built as
\begin{equation}\label{attfno}
\begin{aligned}
\hat{\textbf{u}}:=\operatorname{MLP}(\sigma\left(W \textbf{a} + \mathcal{F}^{-1}(R_{\phi} \cdot \mathcal{F} (\alpha^\mathcal{X}\alpha^\mathcal{C}\textbf{a}))\right)), 
\end{aligned}
\end{equation}
to avoid confusion, we make the input $\textbf{a}$ and output $\textbf{u}$ in bold and omit $\forall x \in \mathcal{D}$. We further introduce the details and modify our model architecture to better learn the high-frequency feature in the next subsection.

\subsection{Model Architecture}
We propose a hierarchical attentive Fourier neural operator combined with convolutional-residual layers and attention mechanisms to learn the function mapping at various resolutions. 

\textbf{Convolutional-Residual Fourier layers:} Inspired by DCNO \cite{anonymous2023dilated}, we propose the convolution-residual Fourier layers which are composed of two main components. In the first component, the input feature is transformed into the frequency domain by the Fast Fourier Transform (FFT) and directly learning an element-wise weight in the frequency domain. We follow the setting of FNO, only reserving lower-frequency components and training the kernel weights on them. This setting aims to learn a smooth mapping to avoid jagged curves in solution spaces. However, this setting may ignore some details about the solution space, especially in solving multiscale PDEs. Since convolution utilizes a much smaller kernel size than the Fourier transform allowing the kernel to capture locally detailed information, we replaced the fully connected residual layers with a convolution layer. This approach pertains to the high-frequency components that are neglected by the Fourier layer.
To prevent confusion, we simplify the coordinates $x$ and denote the input and output of the convolutional-residual Fourier layer at the $k$-th scale as $v^k$ and $\tilde{v}^k$  respectively.
Thus, the Fourier layers could be modulated as:

\begin{equation}
\tilde{v}^k=\sigma\left(\text{Conv}(v^k) +\mathcal{F}^{-1}(R_{\phi}\cdot \mathcal{F}(v^k)\right),
\end{equation}
where $\sigma$ denotes the GELU activation, $R_{\phi}$ represents the kernel weights in the Fourier domain that should be trained.  
Although the convolutional residual layers help to capture some high-frequency features, relying solely on this component is not enough to capture all detailed information. 
We also leverage attention mechanisms to enhance the extraction and integration of information.

\textbf{Attention mechanism:} The attention mechanism can be regarded as a dynamic selection process, where it chooses important features while automatically disregarding irrelevant parts of the input features. The attention mechanism is suitable for learning dependence among pixels in the computer vision field \cite{yuan2020object, geng2021attention}. Generally, the grid data in this work is similar to pixel images. Previous works have conducted thorough experiments to determine the optimal configuration for integrating channel and spatial attention maps in the $\mathbb{R}^d$ scenario. We follow the settings in \cite{woo2018cbam} that serially start with the channel attention.
\begin{equation}
    \begin{gathered}
    \tilde{a}^k_c = \mathcal{A}^{\mathcal{C}}(\tilde{a}^k) \otimes \tilde{a}^k \\
    \tilde{a}^k_{xc} = \mathcal{A}^{\mathcal{X}}(\tilde{a}^k_c)\otimes \tilde{a}^k_c \\
    v^k = \tilde{a}^k_{xc} + \tilde{a}^k
    \end{gathered}
\end{equation}
where $\tilde{a}^k$ serves as the input of the attention layers, which also denotes the downsampled feature of model input $a$. $v^k$ represents the output of the attention layers. $\otimes$ denotes the element-wise product. The $\mathcal{A}^{\mathcal{C}}$ and $\mathcal{A}^{\mathcal{X}}$ denote the channel and spatial attention respectively and have been proved to the equivariant operator before.

\begin{table*}[t]
	\caption{Experiment results on various elliptic equations with various resolutions. $\rightarrow$ denotes the resolution mapping between input and output. For example, $256\rightarrow256$ denotes the input coefficient spaces are $256\times256$ and the output solution spaces are $256\times256$. Performance is measured with mean squared error (MSE with $\times10^{-2}$). The number of parameters and time per epoch are measured for a batch size of 10. For clarity, the best result is in \textbf{bold} and the second best is \underline{underlined}.}
	\label{multiscale elliptic}
	\centering
        
	\begin{small}
		\begin{sc}
			\begin{tabular}{l|c|c|ccccc}
				\toprule
			    \textbf{Methods} & \textbf{Parameters} & \textbf{Time per epoch}   & \multicolumn{2}{c}{\textbf{Trigonometric}} & \multicolumn{2}{c}{\textbf{Darcy rough}} & \textbf{Darcy-Smooth} \\
                & ($\times10^6$) &(s) & $256\rightarrow256$ & $512\rightarrow512$ & $128\rightarrow128$ & $256\rightarrow256$  & $64\rightarrow64$\\
			    \midrule
		      FNO & 2.42 & 7.42       & 1.936 & 1.932 & 2.160 & 2.098 &1.08\\	     
                WMT &   10.32 & 20.03    & 1.043 & 1.087 & 1.573 & 1.621 &0.82\\
                U-NO & 16.39 & 15.42    & 1.256 & 1.245 & 1.368 & 1.332 &1.13\\
                GT &  3.27 &  36.32     & 1.143 & 1.264 & 2.231 & 2.423 &1.70\\
                F-FNO & 4.87 & 10.42     & 1.429 & 1.424 & 1.435 & 1.513 &0.77\\
			HANO & 15.84 & 29.13    & 0.893 & 0.948 & 1.172 & 1.241  &0.79\\
                LSM & 5.81 &  14.26   & \underline{0.832} & \underline{0.782} & \textbf{1.036} & 1.014 &\underline{0.65}\\
                DCNO & 2.27 &  11.73   & 1.056 & 1.209 & 1.276 & \underline{0.948} &0.72\\
                \midrule
                Ours &  5.36 & 10.81 &  \textbf{0.722}  &  \textbf{0.695}  &  \underline{1.161}   &  \textbf{0.904}    &  \textbf{0.59}\\
                \bottomrule
			\end{tabular}
		\end{sc}
	\end{small}
        
\end{table*}

\textbf{Hierarchical architecture:} We attempt to design our model hierarchically, with various scales as inputs. As in multiscale PDEs, multiple scales and regions represent different physical laws \cite{karniadakis2021physics}. The final prediction output is obtained by successively upsampling the outputs in various scales from coarse to fine. Specifically, for the $k$-scale, $\tilde{u}^k$ is concatenated with the interpolation-upsampled $\tilde{u}^{k+1}$ and further projected with a linear layer. More details are presented in the Appendix. 

As the weight matrix is directly parameterized in the Fourier domain, we follow the FNO \cite{li2020fourier} to limit the Fourier series by terminating it at a predefined number of modes. In simple terms, we employ different truncation values at different hierarchical layers to ensure that our model can learn diverse information at different scales. However, large truncation modes would cause computing resources to increase hugely. To balance the computation cost and performance, we set the truncation mode to decrease with the feature scale, as we reckon that large-scale features need more Fourier modes to represent, detailed values are listed in the Appendix.

\subsection{Evaluation metrics}
Previous works \cite{liu2022ht, anonymous2023dilated} proposed H1 loss to solve multiscale PDEs which calculates the loss in the Fourier domain. However, we only use the normalized mean squared error (N-MSE) as the loss function and evaluation metrics, which is defined as
\begin{equation}
    \text{N-MSE}=\frac{1}{B}\sum_{i=1}^B\frac{\|\hat u_i-u_i\|_2}{\|\hat u_i\|_2},
\end{equation}
where $\|\cdot\|_2$ is the 2-norm, $B$ is the batch size, $u$ and $\hat u$ are the network output prediction and ground truth respectively.

\section{Experiments}
\label{sec:experiments}
\textbf{Benchmarks.} We evaluate our method on various PDE benchmarks, including multiscale elliptic equations with various resolutions, Navier-Stokes equations with different viscosity coefficients, and other physics scenarios governed by PDEs. Also, we conduct experiments on the inverse problem of multiscale elliptic equations with noise added to input data.

\textbf{Baselines.} We compare our method with recent and advanced methods. FNO \cite{li2020fourier}, U-NO \cite{rahman2022u}, and F-FNO \cite{tran2021factorized} are FNO-relevant methods that use Fourier transformation to learn the operators directly in the frequency domain. WMT \cite{gupta2021multiwavelet} learns the kernel projection onto fixed multiwavelet polynomial bases. GT \cite{cao2021choose} modify the self-attention to Fourier-type and Galerkin-type attentions with linear complexities to solve the PDEs. HANO \cite{liu2022ht} utilizes hierarchical attention to solve multiscale PDEs. LSM \cite{wu2023solving} solves the PDEs in the latent spectrum domain by decomposing latent features into basic operators.

\subsection{Complex Elliptic Equations}
\label{sec_multi}
The elliptic equation datasets describe the flow of fluid through a porous medium, which are formulated by a second-order linear elliptic equation,
\begin{equation}
    \left\{
	\begin{aligned}
		-\nabla \cdot(a (x) \nabla u (x)) &=f(x), & & x \in D, \\
		u (x) &=0, & & x \in \partial D,
	\end{aligned}
    \right.
\label{eqn:darcy}
\end{equation}
with rough coefficients and Dirichlet boundary. Our model aims to approximate an operator, which maps the input coefficient function to the corresponding output solution space. In contrast to previous works, the coefficient functions show a significant degree of smoothness, leading to correspondingly smooth solutions. We follow the setting in HANO \cite{liu2022ht} and DCNO \cite{anonymous2023dilated} to change the conventional elliptic equations into multiscale cases by modifying coefficients to two-phrase rough ones (\textbf{Darcy-Rough}) or high-contrast trigonometric coefficients (\textbf{Trigonometric}). More details are denoted in the Appendix.


The original experiments of multiscale PDEs using coefficients with resolution $1023\times1023$ to approximate the solution with resolution $256\times256$ or $512\times512$ in the trigonometric setting, which reduce the difficulties as larger inputs might contain more specific information. To enhance the difficulty, we modify the resolution of coefficients to the same as that of the output solution. We also follow the setting in \cite{li2020fourier} and perform the original elliptic equation dataset (\textbf{Darcy-Smooth}) for comparison.

The experiment results are presented in \Cref{multiscale elliptic}, our model achieves the lowest error compared to other operator baselines in most situations, especially in the elliptic equations with trigonometric coefficients. 
Our results indicate that cascade architecture models, such as FNO and DCNO, perform suboptimally in this setting, while hierarchical structures, such as U-NO, HANO, and LSM, perform relatively better. However, the amount of parameters and computations required by HANO and UNO are considerable, our model reduces the parameter quantity by using equivariant attention and convolutions and reaches better prediction performances. To better show the improvement of our model in solving multiscale PDEs, we visualize the prediction solution and error in \Cref{fig4}. 
Compared with FNO, our model achieves lower prediction error, especially in high-frequency components, which shows the improvements of our model to capture high-frequency features.
\subsection{Navier-Stokes Equation}
We consider the 2D Navier-Stokes equation, a standard benchmark proposed in FNO \cite{li2020fourier}, in which the vorticity forms on the two-dimensional torus $\mathbb{T}^2 $.

\begin{figure}[h]
    \centering
    \includegraphics[width=0.98\linewidth]{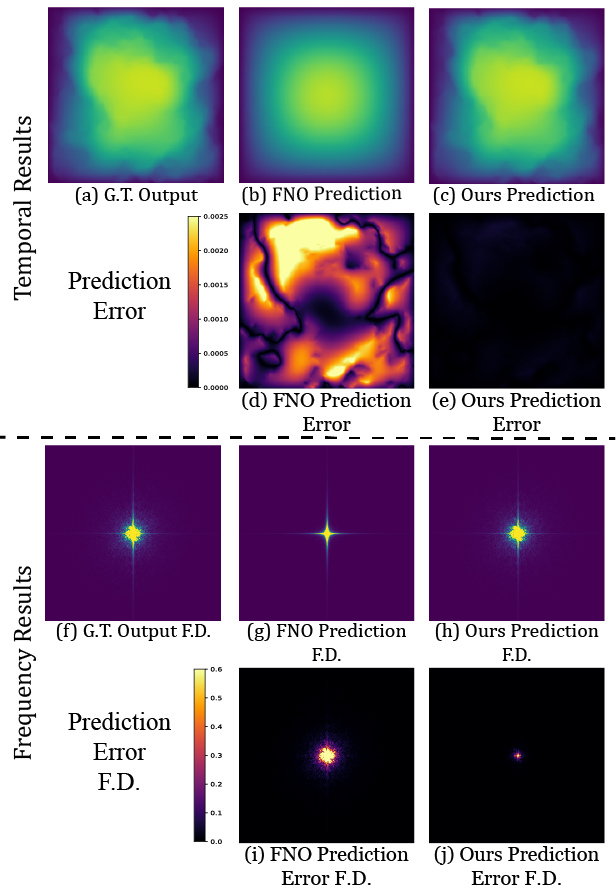}
    \caption{Showcase of Darcy-Rough Elliptic Equations, where the high-frequency components are moved to the center. \textbf{G.T.} and \textbf{F.D.} denote the ground truth and frequency domain. The absolute error is computed as $|u-\hat{u}|$. }
    \label{fig4}
    \vspace{-5pt}
\end{figure}

\begin{table}[t]
	\caption{Experiments on various Navier-Stokes equations and other physical scenarios including  Elasticity, Darcy-Smooth, and Pipe. Performances are measured with MSE ($\times10^{-2}$).  Also, the best result is in \textbf{bold} and the second best is \underline{underlined}.}
	\label{ns}
        
	\centering
                \begin{small}
			\setlength{\tabcolsep}{3pt}
			\begin{tabular}{l|ccc|ccc}
				\toprule
                    \multirow{2}{*}{Methods} & \multicolumn{3}{c}{\textbf{Navier-Stokes}} & \multirow{4}{*}{\textbf{Elasticity}} &  \multirow{4}{*}{\textbf{Pipe}} \\
			         & $\nu=10^{-3}$ & $\nu=10^{-4}$ & $\nu=10^{-5}$ & & & \\
                                                & $T_0=10s$ & $T_0=10s$ & $T_0=10s$ &  &\\
                                                & $T=50s$ & $T=30s$ & $T=20s$ &  \\
			    \midrule
		      FNO   & 1.28 & 8.34 & 19.82  & 5.08   & 0.67\\	     
                WMT   & 1.01 & 11.35 & 15.41  & 5.20  & 0.77\\
                U-NO  & 0.89 & \textbf{5.72} & 17.53  & 4.69   & 1.00\\
                GT    & 1.12 & - & 26.84  & 16.81  & 0.98\\
                F-FNO & 0.92 & 6.02 & 17.98  & 4.72  & 0.59\\
			HANO  & 0.98 & 6.18 & 18.47  & 4.75  & 0.70\\
                LSM   & \underline{0.82} & 6.12 & \underline{15.35}  & \underline{4.08}  & \textbf{0.50}\\
                \midrule
                Ours &  \textbf{0.72}    &  \underline{5.92}  & \textbf{15.09}  & \textbf{3.89}  &  \underline{0.51}   \\ 
                \bottomrule
			\end{tabular}
                \end{small}
                
\end{table}

Specifically, the operator predicts the vorticity after $T_0$ by the input vorticity before $T_0$, the values of $T_0$ and $T$ vary according to the dataset. 
Our experiments consider viscosities with $\nu\in\{10^{-3}, 10^{-4},$
$ 10^{-5}\}$, with smaller viscosities denoting more chaotic flow, are much harder to predict. 
The ‘rollout’ strategy is employed to predict vorticity at each time step in a recursive manner to ensure fair comparisons. 
The final operator could be regarded as approximated by various neural operators. 

\subsection{Other physical scenarios}
We further evaluate our method on Pipe and Elasticity datasets. 

\textbf{Pipe}:
The Pipe dataset focuses on predicting the incompressible flow through a pipe. The input is the pipe structure, while the output is the horizontal fluid velocity within the pipe.
In this dataset, geometrically structured meshes with resolution $129\times129$ are generated. The input and output are the mesh structure and fluid velocity within the pipe.

\textbf{Elasticity}: The Elasticity dataset is designed to predict the internal stress within an incompressible material containing an arbitrary void at its center and an external tensile force is exerted on the material. Originally, the Elasticity data are presented by the point clouds, we follow \cite{wu2023solving} to modify the data into regular grids. The input consists of the material's structural characteristics, while the output represents the internal stress.  These benchmarks estimate the inner stress of incompressible materials with an arbitrary void in the center of the material. In addition, external tension is applied to the material. This benchmark's input and output are the material's structure and inner stress.

We evaluate our model on these datasets to demonstrate the effectiveness of our model in solving general PDE problems. \Cref{ns} summarizes experimental results on Navier-Stokes equations with various coefficients and other physical scenarios. Our model performs better in almost all settings, especially in the Navier-Stokes dataset with small viscosities and Elasticity dataset. Learning high-frequency features may help capture detailed flow changes as flows with smaller viscosities are more chaotic. 

\begin{figure}[t]
    \centering
    \includegraphics[width=0.98\linewidth]{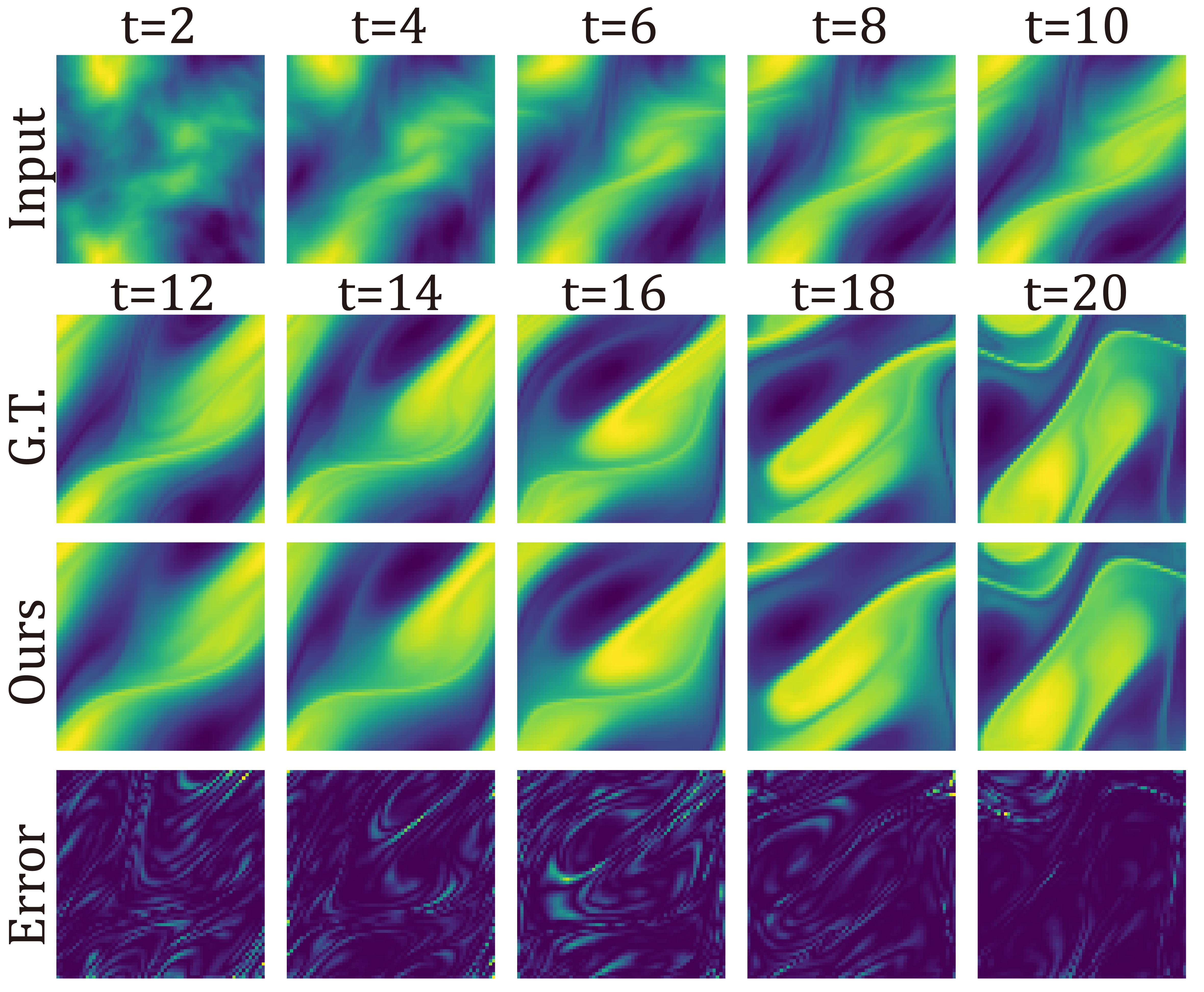}
    \caption{Showcase of prediction results and absolute error of our model in the Navier-Stokes equation dataset. First row: $\textbf{\{2, 4, 6, 8, 10\}}$ time steps input sequence; Second row: $\textbf{\{12, 14, 16, 18, 20\}}$ time steps ground truth sequence; Third row: corresponding prediction sequence; Last row: corresponding absolute prediction error $|\hat{x}-y|$.}
    \label{fig4}
    \vspace{-5pt}
\end{figure}

Our experimental results demonstrate that our model reaches superior performance compared with LSM \cite{wu2023solving}, the previous state-of-the-art approach, which utilizes a combination of self-attention and latent spectral decomposition to solve PDEs within the latent domain. Despite this, it is important to note that LSM is not an operator since it relies on patched multiscale architectures and self-attention mechanisms. We further visualize the results of Navier-Stokes in \Cref{fig4}.

\subsection{Inverse Problems Solving}
In various scientific disciplines such as geological sciences and mathematical derivation, inverse problems are of significant importance. Nonetheless, these problems frequently demonstrate reduced stability compared to their associated forward problems, even when advanced regularization techniques are employed.

Following \cite{anonymous2023dilated}, we evaluate our method for inverse identification problems on multiscale elliptic PDEs. In this experiment, we aim to learn an inverse operator, which maps the solution function space to the corresponding coefficient space $\hat{u}=u+\epsilon N(u)\mapsto a$. Here, $\epsilon$ indicates the extent of Gaussian noise introduced into the training and evaluation data. The noise term $N(u)$ accounts for the sampling distribution and data-related noise.

The experiments about inverse coefficients inference problems on the multiscale elliptic PDEs dataset are presented in \Cref{inverse}. In our experiments, we modify the input and output resolutions to both $256\times256$ in the Darcy Rough and Trigonometric elliptic equations. Since the coefficient function space changes faster than the solution space, this task is more challenging than the forward-solving problem. 

\begin{table}[t]
	\caption{Experiments on inverse coefficient identification tasks. In this experiment, the input solution space and output coefficient space are both $256\times256$. Performances are measured with MSE ($\times10^{-2}$). Also, the best result is in \textbf{bold} and the second best is \underline{underlined}.}
	\label{inverse}
	\centering
            \begin{small}
                \setlength{\tabcolsep}{3pt}
			\begin{tabular}{l|ccc|ccc}
				\toprule
                methods & \multicolumn{3}{c}{Trigonometric} & \multicolumn{3}{c}{Darcy rough}\\
                         & $\epsilon=0$ & $\epsilon=0.01$ & $\epsilon=0.1$ & $\epsilon=0$ & $\epsilon=0.01$ & $\epsilon=0.1$\\
			    \midrule
		      FNO   & 44.74 & 46.34 &  48.43 & 28.41 & 28.98 & 30.65 \\	     
                WMT   & 11.14 & \underline{12.43} &  20.43 & 12.32 & 17.54 & 28.43\\
                U-NO  & 12.97 & 18.54 &  25.87 & 15.64 & 20.54 & 25.34 \\
                GT    & 27.87 & 30.98 &  43.54 & 23.12 &  28.87 & 35.43\\
                F-FNO & 21.46 & 26.98 &  36.34 & 18.73 &  25.23 & 37.54 \\
			HANO  & 9.87 & 13.67 & \underline{20.98} & 8.45 &  \underline{10.43} &  \underline{20.43} \\
                DCNO  & \underline{8.87} & 17.64 & 34.76 & \textbf{6.32} & 11.83 &  23.54 \\
                \midrule
                Ours & \textbf{8.32} & \textbf{10.14} & \textbf{18.24}   & \underline{7.48} & \textbf{9.42} & \textbf{17.32} \\ 
                \bottomrule
			\end{tabular}
            \end{small}
        
\end{table}

The result shows that our model performs well in the inverse coefficient identification problem, which illustrates our model's ability to successfully address the challenges posed by this ill-posed inverse problem with data. Methods such as FNO and F-FNO that learn kernel functions directly in the low-frequency domain have trouble recovering targets with high-frequency components.

\subsection{Ablation study}
To verify the effectiveness of each component in our model, we perform ablation studies in various settings, including removing components ($\textit{w/o}$), replacing them with other components ($\textit{rep}$), and adding some other components ($\textit{add}$).

\begin{itemize}
    \item In the $\textit{w/o}$ part, we consider removing the attention component ($\textit{w/o}$ $\text{Attention}$) or even the Fourier layer ($\textit{w/o}$ $\text{FNO}$).

    \item In the $\textit{rep}$ part, we consider replacing the convolutional-residual layer and attention mechanism with other components and keeping the number of parameters almost unchanged.
    For convolutional-residual layers, we replace them with simple residual layers ($\textit{rep}$ $\text{Conv}{\tiny{\rightarrow}}\text{Res}$) or fully connected layers ($\textit{rep}$ $\text{Conv}{\tiny{\rightarrow}}\text{Fc}$). For the attention mechanism, we replace them with dilation convolution ($\textit{rep}$ $\text{Att}{\tiny{\rightarrow}}\text{d-Conv}$) or multi-layer perceptrons ($\textit{rep}$ $\text{Att}{\tiny{\rightarrow}}\text{MLP}$).

    \item In the $\textit{add}$ part, we add one hierarchical layer with corresponding attention and Fourier layer ($\textit{add}$ $\text{Hier}$). 
\end{itemize}

It is discovered that all components of our model are essential for solving multiscale PDEs after removing experiments. Without the attention and Fourier layer, model performance on all benchmarks will drop seriously. Similar results could be found in replacing experiments, the convolutional-residual Fourier layer and attention can fit the multiscale PDEs benchmarks well. Besides, in the Navier-Stokes dataset, MLP achieved performance comparable to attention, suggesting that previous methods can work well in PDEs where the coefficients do not change rapidly. Additionally, adding a hierarchical layer may improve the performance of the model sometimes, but this will add a significant amount of computation. To balance the efficiency and effectiveness we keep the number of layers of our model. Last but not least, our model performs better than methods with cascade architecture, such as FNO and F-FNO, indicating the importance of solving multiscale PDEs hierarchically. 

\begin{table}[t]
	\caption{Ablation studies on our proposed model, including \textit{removing components (w/o)}, \textit{replacing them with other components (rep)}, and \textit{adding some other components (add)}. Performances are measured with MSE ($\times10^{-2}$) and the best result is in \textbf{bold}.}
	\label{tab:ablation}
	\centering
                \begin{small}
			\begin{tabular}{l|l|ccc}
				\toprule
			\multicolumn{2}{c|}{\multirow{2}{*}{Designs}} & \multicolumn{3}{c}{\textbf{MSE}} \\
                \multicolumn{2}{c|}{}& \scalebox{0.95}{Trigono} & \scalebox{0.95}{Darcy-Rough} & \scalebox{0.95}{Navier-Stokes}\\
                \midrule
                  \multirow{2}{*}{\textit{w/o}} & $\text{Attention}$ & 1.054 & 1.162 & 17.25 \\
                  & $\text{FNO}$ & 1.228 & 1.306 & 22.31  \\
			    \midrule
			     \multirow{4}{*}{\textit{rep}} & $\text{Conv}{\tiny{\rightarrow}}\text{Res}$ & 1.131 & 1.162 & 16.46 \\
                  & $\text{Conv}{\tiny{\rightarrow}}\text{Fc}$ & 1.176 & 1.245 & 15.72  \\
                  & \scalebox{0.95}{$\text{Att}{\tiny{\rightarrow}}\text{d-Conv}$} & 0.894 & 0.954 & 16.12 \\
                  & $\text{Att}{\tiny{\rightarrow}}\text{MLP}$ & 0.985 & 1.034 & 15.63 \\

                  \midrule
                  \textit{add} & $\text{Hier}$ & 0.794 & \textbf{0.901} & 15.36 \\
                  \midrule
                  \multicolumn{2}{c|}{\textbf{Ours}} & \textbf{0.722} & 0.904 & \textbf{15.29} \\
				\bottomrule
  			\end{tabular}
                \end{small}
                
\end{table}

\section{Conclusion}
We propose a novel hierarchical attentive Fourier neural operator that combines convolutional-residual Fourier layers and equivariant attentions for solving multiscale PDEs. Our model utilizes Fourier layers to learn low-frequency features, with convolutional-residual layer and attention mechanisms to capture high-frequency features. Benefits from the above components, our model could capture both local and global information and achieve superior performances in many PDE benchmarks, especially in solving forward and inverse problems of multiscale elliptic PDEs. 

Many works should be done in the future. Researching how to solve more complicated PDEs or even complex dynamics in the real scene and how to use the property of PDEs to design neural network architectures is meaningful. 

{
\small
\bibliography{aaai25}
}

\clearpage
\setcounter{page}{1}

\section{Benchmark Details}
\label{sec:benchmark betails}
We introduce the underlying PDEs of each benchmark and the number of corresponding training and testing samples.

\subsection{Multiscale Elliptic PDEs}
Multiscale elliptic equations are given by second-order linear elliptic equations,
\begin{equation}
    \left\{
	\begin{aligned}
		-\nabla \cdot(a (x) \nabla u (x)) &=f(x) & & x \in D \\
		u (x) &=0 & & x \in \partial D
	\end{aligned}
    \right.
\label{eqn:ms}
\end{equation}
where the coefficient $a(x)\in\left[a_{\min}, a_{\max}\right], \forall x \in D$ and $a_{\min}>0$. 
The coefficient $a(x)$, enables rapid oscillation (for example, with $a(x) = a(x/\varepsilon)$ where $\varepsilon\ll 1$), a significant contrast ratio characterized by $a_{\max}/a_{\min}\gg 1$, and even a continuum of non-separable scales.
 
\subsubsection{Multiscale Trigonometric Coefficient}
We follow the setting in HANO \cite{liu2022ht}, which considers \cref{eqn:ms} with multiscale trigonometric coefficients. The domain $D$ is $[-1,1]^2$, and the coefficient $a(x)$ is defined as $a(x) = \prod \limits_{k=1}^6  (1+\frac{1}{2} \cos(a_k \pi (x_1+x_2)))(1+\frac{1}{2} \sin(a_k \pi (x_2-3x_1)))$, where $a_k$ is uniformly distributed between $2^{k-1}$ and $1.5\times 2^{k-1}$, and the forcing term is fixed as $f(x)\equiv 1$. 
The resolution of the dataset is $1023\times1023$ and lower resolutions are created by downsampling with linear interpolation.

\subsubsection{Two-Phase Coefficient}
The two-phase coefficients and solutions are generated according to FNO \cite{li2020fourier}. 
The coefficients $a(x)$ are generated according to $a \sim \mu:=\psi_{\#} \mathcal{N}\left(0,(-\Delta+c I)^{-2}\right)$ with zero Neumann boundary conditions on the Laplacian. The mapping $\psi: \mathbb{R} \rightarrow \mathbb{R}$ assigns the value $a_{\max}$ to the positive segment of the real line and $a_{\min}$ to the negative segment. The push-forward is explicitly defined on a pointwise basis. The forcing term is fixed as $f(x)\equiv 1$. The solutions $u$ are derived through the application of a second-order finite difference approach on a well-suited grid. The parameters $a_{\max}$ and $a_{\min}$ have the ability to manage the contrast of the coefficient. Additionally, the parameter $c$ regulates the roughness or oscillation of the coefficient; an increased value of $c$ leads to a coefficient featuring rougher two-phase interfaces.

\subsection{Navier-Stokes}
We follow the Navier-Stokes equation in FNO \cite{li2020fourier}. This dataset simulates incompressible and viscous flow on the unit torus, where fluid density is unchangeable. In this situation, energy conservation is independent of mass and momentum conservation.
\begin{equation}
\begin{split}
    \nabla \cdot u &= 0\\
    \frac{\partial w}{\partial t} + u \cdot \nabla w &= \nu \nabla^2 w + f\\
    w|_{t=0}&=w_0,
\end{split}
\end{equation}
where $u$ and $w$ are abbreviated versions of $u(x,t)$ and $w(x,t)$, respectively. $u\in \mathbb{R}^2$  is a velocity vector in 2D field, $w=\nabla \times u$ is the vorticity, $w_0\in \mathbb{R}$ is the initial vorticity  at $t=0$. In this dataset, viscosity  is set as $\nu\in\{10^{-3}, 10^{-4}, 10^{-5}, 10^{-6}\}$ and the resolution of the 2D field is $64\times64$. The number of training and prediction frames is varied in different settings.

\subsection{Elasticity} 
The governing equation of Elasticity materials is:
\begin{equation}\label{equ:solid_pde}
    \rho^{s}\frac{\partial^2 \boldsymbol{u}}{\partial t^2} + \nabla \cdot  \boldsymbol{\sigma}=0,
\end{equation}
where $\rho^s\in\mathbb{R}$ denotes the solid density, $\nabla$ and $\boldsymbol{\sigma}$ denote the nabla operator and the stress tensor respectively. Function $\boldsymbol{u}$ represents the displacement vector of material over time $t$. These benchmarks estimate the inner stress of incompressible materials with an arbitrary void in the center of the material. In addition, external tension is applied to the material. This benchmark's input and output are the material's structure and inner stress.

\subsection{Pipe}
This dataset focuses on the incompressible flow through a pipe. The governing equations are similar to Navier-Stokes equations:
\begin{equation}
    \begin{split}
      \nabla \cdot \boldsymbol{U}&=0\\
      \frac{\partial\boldsymbol{U}}{\partial t} + \boldsymbol{U}\cdot\nabla\boldsymbol{U} &= \boldsymbol{f} - \frac{1}{\rho}\nabla p + \nu\nabla^2\boldsymbol{U}.
    \end{split}
\end{equation}
In this dataset, geometrically structured meshes with resolution $129\times129$ are generated. The input and output are the mesh structure and fluid velocity within the pipe.

We provide details of our benchmarks including the number of training and testing samples and their input solutions in \cref{benchmarks}.

\begin{table}[t]
	\caption{More details about PDEs benchmarks, including the number of training and testing samples with their resolutions. NS is short for Navier-Stokes.}
	\label{benchmarks}
        
	\centering
			\setlength{\tabcolsep}{3pt}
			\begin{tabular}{c|c|c|c}
				\toprule
                Benchmarks & $N_{training}$ & $N_{testing}$ &Resolution \\
			    \midrule
		      Trigonometric   & 1000 & 100 & $512\times512$,$256\times256$ \\
                Darcy-Rough   & 1000 & 100   & $256\times256$,$128\times128$ \\
                Darcy-Smooth & 1000 & 200 & $64\times64$ \\
                \midrule
                NS($\nu=10^{-3}$) & 1000 & 200 & $64\times64$ \\
                NS($\nu=10^{-4}$) & 10000 & 2000 & $64\times64$ \\
                NS($\nu=10^{-5}$) & 1000 & 200 & $64\times64$ \\
                \midrule
                Elasticity & 1000 & 200 & $41\times41$ \\
                Pipe & 1000 & 200 & $129\times129$ \\
                \bottomrule
			\end{tabular}
        
\end{table}

\section{Backgrounds and Proofs }
\subsection{Proofs of  Equivariant of Attentive Convolution}
We follow the definition of \cite{cohen2016group} to define the attentive convolution and reduce the visual self-attention into
\begin{equation}
\begin{aligned}
f_c^{out}(g)=\sum_{\tilde{c}}^{N_{\tilde{c}}} \int_G \alpha_{c, \tilde{c}}(g, \tilde{g}) \psi_{c, \tilde{c}}\left(g^{-1} \tilde{g}\right) f_{\tilde{c}}^{in}(\tilde{g}) \mathrm{d} \tilde{g}.
\end{aligned}
\end{equation}

In this work, we only consider group act on $O(d)$, thus the definition could be further reduced into 

\begin{equation}
\begin{aligned}\label{theorem1proof}
f_c^{out}(x)=\sum_{\tilde{c}}^{N_{\tilde{c}}} \int_{\mathbb{R}^d} \alpha_{c, \tilde{c}}(x, \tilde{x}) \psi_{c, \tilde{c}}\left(x^{-1} \tilde{x}\right) f_{\tilde{c}}^{in}(\tilde{x}) \mathrm{d} \tilde{x}.
\end{aligned}
\end{equation}

Without loss of generality, let $\mathfrak{A}: \mathbb{L}_2(\mathbb{R}^d) \rightarrow \mathbb{L}_2(\mathbb{R}^d)$ denote the attentive group convolution defined by \Cref{theorem1proof}, with $N_{\tilde{c}} = N_{\tilde{c}} =1$, and some $\psi$ which in the following we omit to simplify our derivation. Equivariance of $\mathfrak{A}$ implies that $\forall_{f \in \mathbb{L}_2(\mathbb{R}^d)}, \forall_{\bar{x}, x \in \mathbb{R}^d}$:

\begin{equation}
\begin{gathered}
\mathfrak{A}\left[\mathcal{L}_{\bar{x}}[f]\right](x)=\mathcal{L}_{\bar{x}}[\mathfrak{d}[f]](x) \\
\Leftrightarrow \\
\mathfrak{A}\left[\mathcal{L}_{\bar{x}}[f]\right](x)=\mathfrak{A}[f]\left(\bar{x}^{-1} x\right) \\
\Leftrightarrow \\
\hspace{-3cm}\int_{\mathbb{R}^d} \mathcal{A}\left[\mathcal{L}_{\bar{x}}[f]\right](x, \tilde{x}) \mathcal{L}_{\bar{x}}[f](\tilde{x}) \mathrm{d} \tilde{x}= \\
\hspace{+3cm}\int_{\mathbb{R}^d} \mathcal{A}[f]\left(\bar{x}^{-1} x, \tilde{x}\right) f(\tilde{x}) \mathrm{d} \tilde{x} \\
\Leftrightarrow \\
\hspace{-3cm}\int_{\mathbb{R}^d} \mathcal{A}\left[\mathcal{L}_{\bar{x}}[f]\right](x, \tilde{x}) f\left(\bar{x}^{-1} \tilde{x}\right) \mathrm{d} \tilde{x}= \\
\hspace{+3cm}\quad\int_{\mathbb{R}^d} \mathcal{A}[f]\left(\bar{x}^{-1} x, \bar{x}^{-1} \tilde{x}\right) f\left(\bar{x}^{-1} \tilde{x}\right) \mathrm{d} \tilde{x},
\end{gathered}
\end{equation}

where we once again perform the variable substitution $\tilde{x}\rightarrow\bar{x}^{-1}\tilde{x}$ at the right hand side of the last step. This must hold
for all $f \in \mathbb{L}_2(\mathbb{R}^d)$ and hence:
\begin{equation}\label{thm1}
\forall_{\bar{x} \in \mathbb{R}^d}: \mathcal{A}\left[\mathcal{L}_{\bar{x}} f\right](x, \tilde{x})=\mathcal{A}[f]\left(\bar{x}^{-1} x, \bar{x}^{-1} \tilde{x}\right)
\end{equation}

\subsection{Proof of Symmetry of Fourier transform to $O(d)$}

Let $A\in \mathbb{R}^{d\times d}$ be an invertible matrix, $f:\mathbb{R}^d\to\mathbb{R}$ Lebesgue-integrable and $b\in\mathbb{R}^d$. Consider the function $f_{A,b}:\mathbb{R}^d\to\mathbb{R}$ given by $f_{A,b}(x) = f(Ax+b)$. Then
\[
\mathcal F(f_{(A,b)})(\xi) = \frac{e^{-2\pi i \,\langle A^{-T}\xi, b\rangle}}{|\det A|}\,\mathcal F(f)(A^{-T}\xi)
\]

In particular, if $A$ is an orthogonal matrix, then $|\det A| = 1$ and $A^{-T} = A$, so for all $O\in O(n)$:
\[
\mathcal F(f_{(O,b)})(\xi) =  e^{-2\pi i \,\langle O\xi, b\rangle}\,\mathcal F(f)(O\xi)
\]

We will use the multi-dimensional change of variables formula with the substitution $z = Ax+b$, as well as the identity $\langle \xi, Az\rangle = \langle A^T\xi, z\rangle$.
\begin{equation}
\begin{aligned}
&\mathcal F(f_{(A,b)})(\xi) \\
&= \frac1{(2\pi)^{n/2}} \int_{\mathbb{R}^d} e^{-2\pi i\, \langle\xi, x\rangle} f_{(A,b)}(x)\,dx\\
&=\frac1{(2\pi)^{n/2}\,|\det A|}\int_{\mathbb{R}^d} e^{-2\pi i\,\langle\xi, A^{-1}(Ax+b)\rangle + 2\pi i\,\langle \xi, A^{-1}b\rangle} \\
& f(Ax+b)\,|\det A|dx\\
&= e^{2\pi i \langle \xi, A^{-1}b\rangle}\,\frac1{(2\pi)^{n/2}\,|\det A|}  \int_{\mathbb{R}^d} e^{-2\pi i\,\langle\xi, A^{-1}z\rangle} f(z)\,dz\\
&= \frac{e^{2\pi i \,\langle \xi, A^{-1}b\rangle}}{|\det A|} \,\frac1{(2\pi)^{n/2}}\int_{\mathbb{R}^d} e^{-2\pi i\,\langle A^{-T}\xi, z\rangle}\,f(z)\,dz\\
&= \frac{e^{2\pi i \,\langle A^{-T} \xi, b\rangle}}{|\det A|}\,\mathcal F(f)(A^{-T}\xi).
\end{aligned}
\end{equation}

\section{Model Details}

\subsection{Implementation Details.}
Our model is implemented in PyTorch and conducted on a single NVIDIA A100 40GB GPU. Here are the implementation details of our model.

\subsection{Model Configurations.} Here we present the detailed model configurations of our model in \cref{model config}. In the beginning, the input data will be padded with zeros properly to resolve the division problem in model configurations.

\begin{table}[t]
	\caption{Model configurations.}
	\label{model config}
	\centering
	\begin{small}
		\begin{sc}
			\renewcommand{\multirowsetup}{\centering}
			\setlength{\tabcolsep}{3pt}
			\scalebox{1}{
			\begin{tabular}{c|c|c}
				\toprule
                Model Designs & Hyperparameters & Values \\
			    \midrule 
		      Fourier   & Low-frequency   &  \multirow{2}{*}{$\{24, 12, 6, 3\}$}\\
                Layers    & modes $\{d_{low}^1$, $\cdots$, $d_{low}^K\}$  & \\
                    \midrule 
                              & Channels of each  & \multirow{2}{*}{$\{32, 64, 128, 128\}$} \\
                Hierarchical  & scale $\{d_{\text{c}}^{1},\cdots,d_{\text{c}}^{K}\}$ & \\
                              \cmidrule(lr){2-3}
                Layers        & Number of scales $K$ & $4$ \\
                              \cmidrule(lr){2-3}
                              & Downsmaple ratio $r$ & $2$ \\
                    \midrule
                Training & Learning rate & 0.0005 \\
                Setting  & Batchsize & 10 \\
                \bottomrule
			\end{tabular}}
		\end{sc}
        
	\end{small}
\end{table}

\subsection{Downsample and Upsample Architecture}
In this section, we illustrate the downsampling and upsampling operations in our hierarchical architectures. Our method is similar to that of LSM \cite{wu2023solving}.
\paragraph{Downsampling.} Given deep features $\{\tilde{a}^{k}(x)\}_{{x}\in\mathcal{D}^{k}}$ at the $k$-th scale, The downsampling operation is to aggregate deep features in a local region through maximum pooling and convolution operations, which can be formalized as:
\begin{equation}
\begin{split}
    \{\tilde{a}^{k+1}(x)\}_{{x}\in\mathcal{D}^{k+1}} = \operatorname{Conv}&\Big(\operatorname{MaxPool}\big(\{\tilde{a}^{k}(x)\}_{{x}\in\mathcal{D}^{k}}\big)\Big),\\
    &\text{$k$ from $1$ to $(K-1)$}.
\end{split}
\end{equation}
\paragraph{Upsampling.} Given the features ${{\tilde{u}}^{k+1}(x)}_{x\in\mathcal{D}^{k+1}}$ and ${{\tilde{u}}^{k}(x)}_{x\in\mathcal{D}^{k}}$ corresponding to the $(k+1)$-th and $k$-th scales, respectively, the upsampling procedure involves fusing the interpolated features from the $(k+1)$-th scale and the features from the $k$-th scale using local convolution. This process can be expressed as follows:
\begin{equation}
\begin{split}
    \{{\hat{u}}^{k}(x)\}_{x\in\mathcal{D}^{k}} = 
    &\operatorname{Conv}\Bigg(\operatorname{Concat}\Big(\Big[\operatorname{Interp}\\
    &\big(\{{\tilde{u}}^{k+1}(x)\}_{x\in\mathcal{D}^{k+1}}\big), \{{\tilde{u}}^{k}(x)\}_{x\in\mathcal{D}^{k}}\Big]\Big)\Bigg),\\
    &\text{$k$ from $(K-1)$ to $1$},
\end{split}
\end{equation}
where we adopt the bilinear Interpolation $\operatorname{Inter}(\cdot)$ for 2D data.

\section{More Related Work}
\subsection{Operator Learning}
Suppose $\mathcal{A}$ and $\mathcal{U}$ denote the infinite input and output function spaces. The objective of the operator is to learn a mapping from $\mathcal{A}$ to $\mathcal{U}$ using a finite collection of input and output pairs in the supervised learning way. For any vector function $a\in\mathcal{A}$, $a:\mathcal{D}_{\mathcal{A}}\rightarrow\mathbb{R}^{d_{\mathcal{A}}}$ with $\mathcal{D}_\mathcal{A}\subset\mathbb{R}^d$ and for any vector function $u\in\mathcal{U}$, $u:\mathcal{D}_{\mathcal{U}}\rightarrow\mathbb{R}^{d_{\mathcal{U}}}$, with $\mathcal{D}_\mathcal{U}\subset\mathbb{R}^d$. Given =the training data $\{(a_i, u_i)\}^N_{i=1}$, our objective is to train an operator $G_\theta:\mathcal{A}\rightarrow\mathcal{U}$ which is parameterized by $\theta$, to learn the mapping between input and output function spaces by extracting relationships from $a$ and $u$.

\subsection{Numerical Solvers for Multiscale PDEs} 
In addressing multiscale PDEs, a variety of numerical approaches are available. Numerical homogenization methods \cite{engquist2008asymptotic} aim to create a finite-dimensional approximation space for solution exploration. Fast solvers like multilevel and multigrid methods \cite{hackbusch2013multi, xu2017algebraic} can be considered as an extension of numerical homogenization. Recently, operator-adapted wavelet methods, such as Gamblets \cite{owhadi2017multigrid}, have been developed to solve linear PDEs with rough coefficients, representing a progression beyond numerical homogenization. Nevertheless, handling multiscale PDEs poses inherent challenges for numerical methods, given that the computational cost tends to scale inversely proportional to the finest scale $\epsilon$ of the problem. In recent years, there has been increasing exploration of neural network methods for solving multiscale PDEs \cite{anonymous2023dilated, liu2022ht}.

\subsection{More FNO Related Work} 
Networks inspired by FNO have been verified in various domains, including computer vision and time series forecasting \cite{ovadia2023vito, ovadia2023ditto}. AFNO \cite{guibas2021adaptive} leverages kernel in the Fourier domain as a token mixer within the transformer, aiming at reducing computational complexity and enhancing performance in segmentation tasks. FEDformer \cite{zhou2022fedformer} harnesses sparse basic elements in the Fourier frequency domain to create a frequency-enhanced transformer. Meanwhile, GFNet \cite{rao2021global} employs the element-wise multiplication of learnable global filters with features in the frequency domain to improve the performance in classification and transfer learning tasks.

\subsection{Visual Attention Methods}
The attention mechanism can be regarded as a process of adaptive selection based on input features. It has yielded advantages in numerous visual tasks, including image classification \cite{woo2018cbam}, object detection \cite{dai2017deformable, hu2018relation}, and semantic segmentation \cite{yuan2020object, geng2021attention}. In computer vision, the attention mechanism is usually be divided into three main categories: channel attention, spatial attention, and temporal attention. For instance, SENet \cite{hu2018squeeze} utilizes global average pooling on the channel dimension to modulate the corresponding channel attention. Complementary channel attention akin to that of CBAM \cite{woo2018cbam} and BAM \cite{park2018bam} utilize similar strategies for spatial attention and combine spatial and channel attention in series and parallel respectively. Recent research in visual attention aims to integrate the strengths of various attention mechanisms to create more holistic attention \cite{hu2018squeeze, guo2022attention}.

\section{More Visualization Results}
We visualize more results on the Trigonometric dataset compared to FNO.

\begin{figure*}[t]
    \centering
    \includegraphics[width=0.98\linewidth]{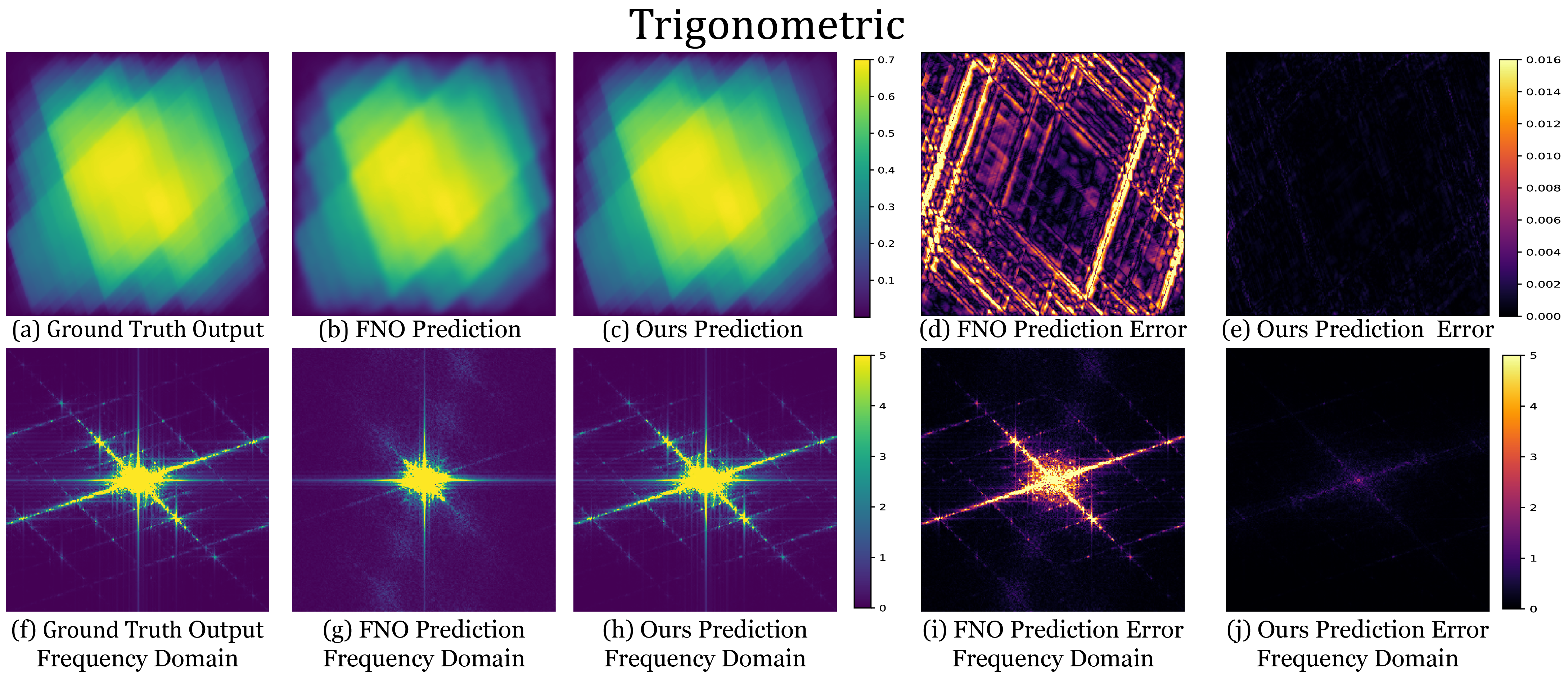}
    \caption{Showcase of multiscale PDEs, where the high-frequency components are moved to the center. \textbf{G.T.} and \textbf{F.D.} denote the ground truth and frequency domain. The absolute error is computed as $|u-\hat{u}|$. }
    \label{fig4}
\end{figure*}

\end{document}